\documentclass[letterpaper]{article}
\usepackage{aaai}
\usepackage{times}
\usepackage{helvet}
\usepackage{courier}
\usepackage{adjustbox}
\usepackage{subcaption}
\usepackage{stfloats}
\usepackage{pifont}
\usepackage{flushend}
\usepackage{balance}
\usepackage[utf8]{inputenc}
\newcommand{\cmark}{\ding{51}}%
\newcommand{\xmark}{\ding{55}}%
\usepackage[maxbibnames=99]{biblatex}
\begin{document}
%
\title{LiftFormer: 3D Human Pose Estimation using attention models}
\author{Adrian Llopart\\
Kirin AI\\
Huawei Technologies UK, London\\
}
\maketitle
\begin{abstract}
\begin{quote}
Estimating the 3D position of human joints has become a widely researched topic in the last years. Special emphasis has gone into defining novel methods that extrapolate 2-dimensional data (keypoints) into 3D, namely predicting the root-relative coordinates of joints associated to human skeletons.
The latest research trends have proven that the Transformer Encoder blocks aggregate temporal information significantly better than previous approaches. Thus, we propose the usage of these models to obtain more accurate 3D predictions by leveraging temporal information using attention mechanisms on ordered sequences human poses in videos.

Our method consistently outperforms the previous best results from the literature when using both 2D keypoint predictors by 0.3 mm (44.8 MPJPE, 0.7\% improvement) and ground truth inputs by 2mm (MPJPE: 31.9, 8.4\% improvement) on Human3.6M. It also achieves state-of-the-art performance on the HumanEva-I dataset with 10.5 P-MPJPE (22.2 \% reduction). The number of parameters in our model is easily tunable and is smaller (9.5M) than current methodologies (16.95M and 11.25M) whilst still having better performance.
Thus, our 3D lifting model's accuracy exceeds that of other end-to-end or SMPL approaches; and is comparable to many multi-view methods. 

\end{quote}
\end{abstract}

\section{Introduction}
Human pose estimation from monocular images is an important research avenue that has gained much momentum lately. The reason behind this is that human pose estimation allows for better gesture/action recognition, game/device control and more precise 3D avatar generation, among other applications. However, currently, estimating precisely the 3D pose of humans requires multi-camera viewpoints, depth sensors or tracking motion capture suits, none of which are available to the general public. Human pose estimation started by predicting sets of 2D joints given a monocular image (either in a top-down or bottom-up fashion), but were soon replaced by architectures that estimated the 3D coordinates (either absolute coordinates w.r.t the camera frame or relative coordinates to the pelvis joint, also known as \textit{root}). Other methods like SMPL try to parameterize the full human body at a higher computational cost.

\begin{figure}[t]
    \centering
    \includegraphics[width=0.5\columnwidth]{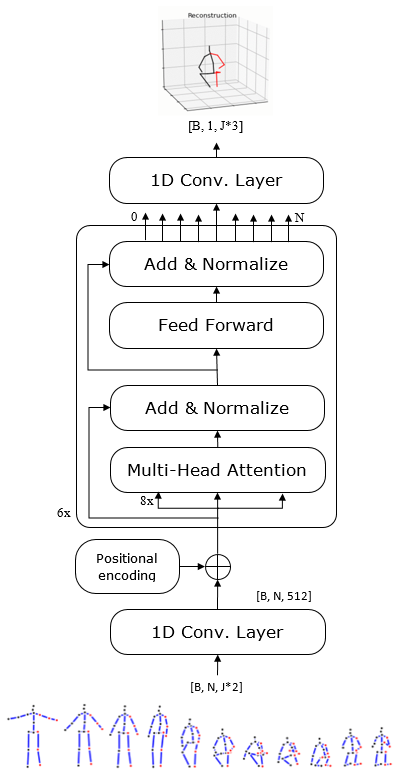}
    \caption{Our Transformer Encoder model takes a sequence of 2D keypoints (bottom) as input and generates 3D pose estimates (top) using self-attention on long-term information.}
    \label{fig:architecture}
\end{figure}

This work focuses on the estimation of 3D keypoints, corresponding to the most notable joints in the human body, from 2D keypoints. This is commonly known in the literature as \textit{lifting from 2D to 3D}. Even though this corresponds to only a smaller part of the problem of keypoint estimation, there are inherently multiple 3D poses that can be represented by a set of 2D coordinates. Previous works \cite{moon19, chang20} have tackled this problem by predicting the absolute coordinates w.r.t the camera. However, the majority of the research in this field  still deals with \textit{root-relative} estimations, that is, predicting the joint coordinates w.r.t the \textit{pelvis} joint \cite{martinez17, hossain18, lee18, dabral18, zhao19, pavllo19, liu20}.  

Additionally, the literature proves that leveraging temporal information results in substantial performance improvements. Whilst early approaches employed the use of Recurrent Neural Networks (RNNs), these methods were soon overshadowed by fully convolutional architectures \cite{pavllo19, cheng19, cheng20}. Lately, the research has shifted to the introduction of \textit{attention} models \cite{liu20}. This work argues that adding attention mechanisms enhances the models temporal coherency by exploiting long-range temporal relationships across frames/poses. 

An example of another well established attention-based model is the Transformer \cite{vaswani17} and its variants \cite{dai19} \cite{nitaev20}, which have been used in a wide array of tasks, not only in Natural Language Processing (NLP), where it was first introduced, but also in the field of speech recognition \cite{zhang20}, object detection \cite{carion20} and image inpainting \cite{chen20}, among others. Furthermore, the ALBERT paper \cite{lan19} shows how to reduce the model size of the Transformer, whilst maintaining the same performance, by sharing weights in the multi-attention heads.

The successful implementation in so many diverse tasks of the Transformer, as a self-attention model, sparks the question of how well it would perform in a \textit{lifting} scenario. Similarly to \cite{pavllo19} and \cite{liu20}, the Transformer Encoder architecture can be used to upscale the dimensionality of sequences of keypoints via its own self-attention. Hence, this paper proposes the Liftformer model, which leverages this type of attention model to predict 3D temporally cohesive joint coordinates from 2D input keypoints. The LiftFormer model not only surpasses all previous \textit{lifting} literature, but does so with fewer parameters.

\section{Related work}
Human pose estimation, from both single images and videos, has been researched   largely in the last years. Originally, most work was based on pre-engineered models, with a large number of constraints, to solve the high degree of dependencies among human joints \cite{ionescu13, ionescu14, amin13, ramakrishna12}. With the development of Convolutional Neural Networks (CNNs), novel pose estimators were created which reconstructed 3D poses end-to-end directly from RGB images or intermediate 2D predictions. This research rapidly surpassed the accuracy of previous hand-made estimators.
\\
\subsubsection{End-to-end models}
Many works have pursued the usage of features extracted from RGB images as a way to obtain 3D \textit{root-relative} pose predictions. Tekin \textit{et al.} \cite{tekin17} leverages both 2D heatmaps and image features by building a two-stream network which aggregates both features through a trainable fusion strategy. Sun \textit{et al. } \cite{sun17} argues that regression methods do not take advantage of the actual structure of human skeletons. For this reason they propose working with \textit{bones}, instead of \textit{joints}, and exploiting the joint connection structure by defining a \textit{compositional loss function}. Pavlakos \textit{et al.} \cite{pavlakos18} enhances datasets and trained joint regressors using ordinal depth relations. Luvizon \textit{et al.} \cite{luvizon18} propose a multitask framework for simultaneous 2D/3D pose estimation and action recognition. To achieve pose estimation they use a differentiable soft-argmax for volumetric heatmaps, which has now become one of the most common ways to solve this task. Zhou \textit{et al.} \cite{zhou19} use Part-Centric Heatmap Triplets to represent the relative depth information of end-joints. That is, three polarized heatmaps, corresponding to the different state of the local depth ordering of the part-centric joint pairs. Zhao \textit{et al.} \cite{zhao19} employ Graph convolutions that take as input the concatenation of pooled features from the backbone of a 2D keypoint predictor and the predictions themselves. However, the accuracy of their results is not that good compared to other state-of-the-art (SOTA) research. Yang \textit{et al.} \cite{yang18} add adversarial learning to a SOTA 3D detector to discriminate humanly plausible poses, which are then aggregated to visual features for Action Recognition.

Moon \textit{et al.} \cite{moon19} is one of the few papers that deal with absolute 3D coordinates, that is, coordinates relative to the camera itself and not to the \textit{root} (i.e subject's pelvis). They employ two networks: the former predicts the 3D root relative coordinates by obtaining volumetric heatmaps (introduced by \cite{pavlakos17}) and applying a 3D \textit{soft-argmax}. The latter predicts a correction factor that multiplies a pre-defined depth distance based on the size of the human within the image and the \textit{intrinsic} camera parameters. 

\subsubsection{Lifitng: Two step pose estimation}
Formally, \textit{two-step} pose estimation models break down the complexity of 3D pose estimation by dividing the problem into two steps: \textbf{i.} They employ common 2D pose estimators to obtain accurate 2D poses of humans in images, usually in a top-down fashion by first predicting bounding boxes around each human instance and then procesing only those regions of interest. \textbf{ii.} They then \textit{lift} these joints to 3D by predicting their depth relative to the root (pelvis).

Martinez \textit{et al.} \cite{martinez17} were the first to introduce the concept of \textit{lifting 2D to 3D}, where they proved that using simple and computationally inexpensive models (a residual block architecture) one is able to accurately predict the depth of previously predicted 2D keypoints. Another idea was propsed by Fang \textit{et al.} \cite{fang18} who introduced grammar models and the usage of Bi-RNNs to explicitly incorporate a set of knowledge regarding human body configuration. The RNNS don't model temporal data, but instead model a chain-like grammar for single images. Similarly to \cite{moon19} and \cite{martinez17}, Chang \textit{et al.} \cite{chang20} propose \textit{PoseLifter}, a method to lift 2D to absolute 3D coordinates using residual blocks.
\\
\subsubsection{Video pose estimation}
In the last years, given that \textit{lifting} approaches have performed so well in the past, there has been a significant and successful trend to aggregate multiple poses to produce more temporally cohesive predictions. 

Hossain \textit{et al.} \cite{hossain18} define their pose lifter as a sequence-to-sequence model composed of layer-normalized LSTM units with shortcut connections. Similarly, Lee \textit{et al.} \cite{lee18} used a Stacked Hourglass \cite{newell16} to predict 2D poses which are then temporally aggregated using p-LSTMs. In contrast, Dabral et el. \cite{dabral18} temporally aggregated 2D poses, predicted with standard 2D keypoint estimators, by using a simple Fully Connected Network (FCN) with ReLU activations.

Pavllo \textit{et al.} \cite{pavllo19} use Temporal Convolutions Networks (TCN) to lift 2D keypoints to 3D. The 2D keypoints are obtained from common object detectors, like Stacked Hourglass (SH), Mask-RCNN \cite{he17} and Cascaded Pyramid Network (CPN) \cite{chen18} pretrained on COCO \cite{lin14} and then finetuned on \textit{Humans3.6M} \cite{ionescu13}. They prove that using ground truth 2D predictions as inputs produce the best result, followed by CPN detections, MRCNN and, finally, SH. They also investigate how larger receptive fields achieve higher accuracy.

This is also demonstrated in Liu \textit{et al.} \cite{liu20}. They are also amongst the first to introduce the concept of \textit{attention} into the 3D keypoint estimation field. They employ attention by dividing their model into \textit{Temporal attention}, which calculates the positive cosine similarity between 2D joints, and \textit{kernel attention}, which is a combination of TCN and Linear projection units. They prove how attention is beneficial for lifting from 2D to 3D. 

Other approaches work by leveraging the original image features when \textit{lifting}. Cheng \textit{et al.} \cite{cheng19} build complex models that focus on occlusions within the time domain. They employ MRCNN to detect humans and SH to detect the 2D keypoints. They clearly state that \textit{"by using incomplete 2D keypoints, instead of complete but incorrect ones, our networks are less affected by the error-prone estimations of occluded keypoints"}. For this reason, they refine their 2D predictions with optical flow followed by a dilated 2D TCN, which removes occlusions. Likewise, the occluded 3D ground truths are removed by applying a \textit{Cylindrical Man model}. Finally, the refined 2D keypoints are passed through a 3D TCN and matched with the modified 3D GTs. Their models achieve SOTA but at the price of a very convoluted system. 

Cheng \textit{et al.} \cite{cheng20} improves the state-of-the-art by using a discriminator to assess if the generated poses are valid. Specifically, they use the Kinematic Chain Space (KCS) model, defined in \cite{wandt19}, and expand it temporally (TKCS). They also use multi-scale 2D heatmaps, instead of keypoints directly, extracted from HRNet \cite{sun19}. By embedding the differently scaled features into 3D keypoints, KCS and TKCS, and then passing the results into a trained discriminator, they are able to achieve SOTA.
\\
\subsubsection{Attention models}
Vaswani \textit{et al.} \cite{vaswani17} demonstrated how attention-based \cite{bahdanau14} models, such as the Transformer, can achieve state-of-the-art results in machine translation tasks. These attention methods aggregate information from an entire input sequence into each of its elements. Their computational efficiency and improved long-term memory makes them better mechanisms at dealing with long input sequences than the previous RNNs or LSTMs. For this reason, Transformers have now been widely used in many NLP tasks \cite{devlin18, lan19}, but have also been present in other fields like speech recognition (Transformer Transducer \cite{zhang20}), object detection (DETR \cite{carion20}) and image inpainting (Image GPT \cite{chen20}), among others.

\section{LiftFormer architecture}
We propose the usage of a Transformer architecture for \textit{lifting}. To this end, we employ the \textit{base} Transformer Encoder presented in \cite{vaswani17} as our baseline, that is 512 dimension hidden layers, 8 multi-attention heads and 6 encoder blocks. The general idea is to pass 2D keypoint sequences through several Transformer Encoder blocks to generate one 3D pose prediction, corresponding to the center of the input sequence/receptive field, and then repeat the process for all timesteps in a sliding window fashion. Thus, during training, the model leverages temporal data from past and future frames to be able to create temporally cohesive predictions.

As shown in Figure \ref{fig:architecture}, certain modifications must take place so that the input and output dimensionality match. Namely, the input to the Transformer Encoder is reprojected from an input dimension of [N, 34] to [N, 512], where \textit{N} is the receptive field and 34 corresponds to 17 joints times 2 (number of coordinates). The reason for this is so that the input dimensionality matches that of the hidden layers inside the Transformer Encoder block. These input sequences can either come from 2D predictors, which estimate the 2D poses from frames, or directly from the datasets 2D ground truths. Hence, these predictions are in the image space.

Then, temporal encoding is aggregated by adding a vector to the input embeddings. This is so that the model can make use of the temporal order in the sequence. In the original Transformer paper \cite{vaswani17}, this is referred to as \textit{positional encoding} which is used to inject some information about the relative or absolute position of the tokens in the sequence. We employ the same idea, using sine and cosine functions to create these \textit{temporal} embeddings which are then summed to the re-projected inputs. The injected \textit{temporal} embeddings must therefore have the same dimensionality as the inputs. The resulting features are then sent to the Transformer Encoder, which processes them.

Since the Decoder part of the Transformer is not being used, and due to the residual connections within the self-attention, the dimension of the output will be exactly the same as the input, that is [N, 512]. Similarly to the BERT model \cite{devlin18} during classification tasks, we choose the output token corresponding to the middle of the receptive field, which corresponds to the pose to be \textit{lifted} within the input sequence. This is because during training, half of the receptive field corresponds to \textit{past} poses, and the other half to \textit{future} poses. The center pose within the receptive field is exactly the pose we are trying to \textit{lift}.

Finally, we reproject the output embeddings into the desired dimensionality, from [1, 512] to [1, 51] using again a 1D convolutional layer, where 51 corresponds to 17 joints times 3 (\textit{x}, \textit{y} and \textit{z} coordinates). The loss is then calculated using MPJPE against the datasets 3D ground truths and the error is back-propagated

During inference, the model architecture could stay the same or work in a \textit{causal} fashion, where only \textit{past} frames will be used in the receptive field. Hence, the model will work in a sliding window fashion so that all frames get their 3D reconstructions.

One of the benefits of this architecture is that you can modify the receptive field and the number of multi-attention heads, without affecting the models size. Additionally, several ablation studies have been performed which show how modifying the Transformer's hyper-parameters change model size and accuracy results (Section \ref{ablations}).

\begin{table*}[tp]
\begin{adjustbox}{width=1\textwidth}
\begin{tabular}{l|ccccccccccccccc|c}
& Dir. & Disc. & Eat & Greet & Phone & Photo & Pose & Purch. & Sit & SitD. & Smoke & Wait & WalkD. & Walk & WalkT. & Avg \\
\hline
Martinez \textit{et al.} ICCV’17 \cite{martinez17} & 51.8 & 56.2 & 58.1 & 59.0 & 69.5 & 78.4 & 55.2 & 58.1 & 74.0 & 94.6 & 62.3 & 59.1 & 65.1 & 49.5 & 52.4 & 62.9 \\
Hossain \textit{et al.} ECCV’18 \cite{hossain18} & 48.4 & 50.7 & 57.2 & 55.2 & 63.1 & 72.6 & 53.0 & 51.7 & 66.1 & 80.9 & 59.0 & 57.3 & 62.4 & 46.6 & 49.6 & 58.3\\
Zhao \textit{et al.} CVPR’19 \cite{zhao19} & 47.3 & 60.7 & 51.4 & 60.5 & 61.1 & 49.9 & 47.3 & 68.1 & 86.2 & \textbf{55.0} & 67.8 & 61.0 & 60.6 & 42.1 & 45.3 & 57.6\\
Luvizon \textit{et al.} CVPR’18 \cite{luvizon18} & 49.2 & 51.6 & 47.6 & 50.5 & 51.8 & 60.3 & 48.5 & 51.7 & 61.5 & 70.9 & 53.7 & 48.9 & 57.9 & 44.4 & 48.9 & 53.2\\
Chang \textit{et al.} \cite{chang20} & 44.8 & 48.2 & 48.5 & 51.5 & 54.5 & 64.4 & 47.9 & 47.8 & 60.7 & 76.4 & 52.5 & 50.8 & 55.3 & 39.0 & 42.2 & 52.5\\
Lee \textit{et al.} ECCV’18  \cite{lee18} & \textbf{40.2} & 49.2 & 47.8 & 52.6 & 50.1 & 75.0 & 50.2 & 43.0 & 55.8 & 73.9 & 54.1 & 55.6 & 58.2 & 43.3 & 43.3 & 52.8\\
Dabral \textit{et al.} ECCV’18 \cite{dabral18} & 44.8 & 50.4 & 44.7 & 49.0 & 52.9 & 61.4 & 43.5 & 45.5 & 63.1 & 87.3 & 51.7 & 48.5 & 52.2 & 37.6 & 41.9 & 52.1\\
Pavllo \textit{et al.} CVPR’19 \cite{pavllo19}& 45.2 & 46.7 & 43.3 & 45.6 & 48.1 & 55.1 & 44.6 & 44.3 & 57.3 & 65.8 & 47.1 & 44.0 & 49.0 & 32.8 & 33.9 & 46.8\\
Liu \textit{et al.} CVPR’20 \cite{liu20} & \underline{41.8} & \underline{44.8} & \textbf{41.1} & 44.9 & \underline{47.4} & \underline{54.1} & \underline{43.4} & \underline{42.2} & 56.2 & 63.6 & \underline{45.3} & \underline{43.5} & \textbf{45.3} & \textbf{31.3} & \underline{32.2} & \underline{45.1} \\
\hline
Liftformer (n=27 CPN) & 45.7 & 49.2 & 47.1 & 47.2 & 50.72 & 57.7 & 46.3 & 44.5 & 58.2 & 64.9 & 49.3 & 45.6 & 50.0 & 35.8 & 37.2 & 48.6\\
Liftformer (n=81 CPN) & 43.8. & 46.4 & 44.1 & \underline{44.6} & 48.0 & 54.9 & 43.7 & 42.5 & \textbf{54.9} & 62.9 & 45.8 & 44.1 & \underline{47.3} & 33.1 & 34.0 & 46.0\\
Liftformer (n=243 CPN)  & 42.2 & \textbf{44.5} & \underline{42.6} & \textbf{43.0} & \textbf{46.9} & \textbf{53.9} & \textbf{42.5} & \textbf{41.7} & \underline{55.2} & \underline{62.3} & \textbf{44.9} & \textbf{42.9} & \textbf{45.3} & \underline{31.8} & \textbf{31.8} & \textbf{44.8}\\
\end{tabular}
\end{adjustbox}

\caption{\textbf{Protocol 1} with MPJPE (mm): Reconstruction error on Human3.6M. Input 2D joints are acquired by detection. CPN - Cascaded Pyramid Network.}
\label{h36m:mpjpeCPN}
\end{table*}

\begin{table*}[tp]
\begin{adjustbox}{width=1\textwidth}
\begin{tabular}{l|ccccccccccccccc|c}
& Dir. & Disc. & Eat & Greet & Phone & Photo & Pose & Purch. & Sit & SitD. & Smoke & Wait & WalkD. & Walk & WalkT. & Avg \\
\hline
Wandt \textit{et al.} CVPR'19 \cite{wandt19} & 50.0 & 53.5 & 44.7 & 51.6 & 49.0 & 58.7 & 48.8 & 51.3 & 51.1 & 66.0 & 46.6 & 50.6 & 42.5 & 38.8 & 60.4 & 50.9 \\
Martinez \textit{et al.} ICCV’17 \cite{martinez17} & 37.7 & 44.4 & 40.3 & 42.1 & 48.2 & 54.9 & 44.4 & 42.1 & 54.6 & 58.0 & 45.1 & 46.4 & 47.6 & 36.4 & 40.4 & 45.5 \\
Zhao \textit{et al.} CVPR’19 \cite{zhao19} & 37.8 & 49.4 & 37.6 & 40.9 & 45.1 & 41.4 & 40.1 & 48.3 & 50.1 & 42.2 & 53.5 & 44.3 & 40.5 & 47.3 & 39.0 & 43.8 \\
Hossain \textit{et al.} ECCV’18 \cite{hossain18} & 35.2 & 40.8 & 37.2 & 37.4 & 43.2 & 44.0 & 38.9 & 35.6 & 42.3 & 44.6 & 39.7 & 39.7 & 40.2 & 32.8 & 35.5 & 39.2 \\
Lee \textit{et al.} ECCV’18 \cite{lee18} & \underline{32.1} & 36.6 & 34.4 & 37.8 & 44.5 & 49.9 & 40.9 & 36.2 & 44.1 & 45.6 & 35.3 & 35.9 & 37.6 & 30.3 & 35.5 & 38.4 \\
Pavllo \textit{et al.} CVPR’19 \cite{pavllo19} & 35.2 & 40.2 & 32.7 & 35.7 & 38.2 & 45.5 & 40.6 & 36.1 & 48.8 & 47.3 & 37.8 & 39.7 & 38.7 & 27.8 & 29.5 & 37.8 \\
Liu \textit{et al.} CVPR’20 \cite{liu20} & 34.5 & 37.1 & 33.6 & 34.2 & \underline{32.9} & 37.1 & 39.6 & 35.8 & 40.7 & 41.4 & 33.0 & 33.8 & 33.0 & 26.6 & 26.9 & 34.7 \\
\hline
Liftformer (n=27 GT) & 37.5 & 41.0 & 36.2 & 35.9 & 37.5 & 40.3 & 41.8 & 34.8 & 42.5 & 43.5 & 37.2 & 38.5 & 35.7 & 30.6 & 31.6 & 37.7 \\
Liftformer (n=81 GT)  & 34.2 & 37.6 & 31.5 & 33.1 & 34.8 & 37.5 & 37.6 & 33.0 & 41.5 & 43.0 & 34.2 & 33.8 & 32.8 & 26.2 & 27.3 & 34.5 \\
Liftformer (n=243 GT) & 33.2 & \underline{35.3} & \textbf{29.8} & \underline{31.1} & \underline{32.9} & \underline{35.1} & \underline{35.5} & \underline{31.5} & \underline{37.2} & \textbf{38.1} & \underline{32.6} & \underline{33.1} & \underline{30.9} & \textbf{24.3} & \underline{26.1} & \underline{32.5}\\
\hline
Liftformer (n=27 GT + WS) & 37.3 & 41.4 & 35.9 & 35.8 & 37.6 & 40.7 & 41.3 & 36.0 & 41.8 & 42.9 & 37.2 & 38.8 & 36.5 & 30.9 & 31.6 & 37.7\\
Liftformer (n=81 GT + WS) & 34.3 & 38.7 & 33.2 & 33.4 & 35.3 & 38.4 & 38.8 & 32.9 & 41.7 & 43.4 & 35.1 & 35.7 & 33.2 & 27.4 & 28.1 & 35.3\\
Liftformer (n=243 GT + WS) & \textbf{31.8} & \textbf{34.6} & \underline{30.1} & \textbf{30.4} & \textbf{32.4} & \textbf{34.3} & \textbf{34.6} & \textbf{31.1} & \textbf{36.9} & \underline{38.9} & \textbf{32.2} & \textbf{32.1} & \textbf{30.4} & \textbf{24.4} & \underline{24.9} & \textbf{31.9}\\
\end{tabular}
\end{adjustbox}

\caption{\textbf{Protocol 1} with MPJPE (mm): Reconstruction error on Human3.6M. Input 2D joints from ground truth. GT - Ground truth. WS - Weight sharing between attention layers}
\label{h36m:mpjpeGT}
\end{table*}

\subsection{Weight sharing}
Lan \textit{et al.} \cite{lan19} propose sharing parameters across layers to build more efficient models whilst maintaining the accuracy. They show how sharing either all parameters in each Encoder block or only the FFN parameters, will considerably hurt the overall performance. On the other hand, sharing only the attention layers parameters seems to barely affect the final accuracy and, in some cases, even improve it. In all cases, the total number of parameters is considerably decreased. More information is shown in the ablation studies in Table \ref{ablat:weight_sharing}.

\section{Experiments}
\subsection{Datasets and Evaluation Protocols}
We evaluate on two motion capture datasets: Human3.6M \cite{ionescu13} and HumanEva \cite{sigal10} , which have been commonly used in the literature. For both datasets we follow the evaluation procedure of previous works \cite{lee18, cheng19, pavllo19, zhou19, liu20, cheng20}, among others.

Human3.6M consists of 3.6 million frames of 11 different subjects, however only 7 of them are annotated. The subjects performed up to 15 different types of actions, which were recorded from 4 different viewpoints. We train on 17 joints with subjects S1, S5, S6, S7 and S8. The evaluation is done on S9 and S11.

HumanEva is a much smaller dataset in comparison, with only 3 subjects and recorded from 3 viewpoints. We evaluate on 2 actions (\textit{Walk}, \textit{Box}) and report all our results for models trained on all actions and 15 joints.

In our experiments, we consider three evaluation protocols, all of which are relative to the \textit{root}: \textbf{Protocol 1} is the mean Euclidean distance between predicted root-relative joint positions and ground truth joint positions (MPJPE in mm). \textbf{Protocol 2} is the error after applying a similarity transformation (Procrustes alignment) with the ground truth in translation, rotation and scale (P-MPJPE in mm). \textbf{Protocol 3} is the mean per joint velocity, which evaluates the smoothness and stability of predictions over time (MPJVE in mm/s).

\begin{table*}[tp]
\begin{adjustbox}{width=1\textwidth}
\begin{tabular}{l|ccccccccccccccc|c}
& Dir. & Disc. & Eat & Greet & Phone & Photo & Pose & Purch. & Sit & SitD. & Smoke & Wait & WalkD. & Walk & WalkT. & Avg \\
\hline
Martinez \textit{et al.} ICCV’17 \cite{martinez17} & 39.5 & 43.2 & 46.4 & 47.0 & 51.0 & 56.0 & 41.4 & 40.6 & 56.5 & 69.4 & 49.2 & 45.0 & 49.5 & 38.0 & 43.1 & 47.7\\
Hossain \textit{et al.} ECCV’18 \cite{hossain18} & 35.7 & 39.3 & 44.6 & 43.0 & 47.2 & 54.0 & 38.3 & 37.5 & 51.6 & 61.3 & 46.5 & 41.4 & 47.3 & 34.2 & 39.4 & 44.1\\
Wandt \textit{et al.} CVPR'19 \cite{wandt19} & 33.6 & 38.8 & 32.6 & 37.5 & 36.0 & 44.1 & 37.8 & 34.9 & 39.2 & 52.0 & 37.5 & 39.8 & 34.1 & 40.3 & 34.9 & 38.2\\
Chang \textit{et al.} \cite{chang20} & 32.1 & 34.9 & 43.4 & 36.9 & 35.4 & 35.1 & 30.8 & 34.3 & 57.3 & 40.4 & 44.9 & 35.1 & 24.9 & 46.6 & 30.0 & 37.7\\
Dabral \textit{et al.} ECCV’18 \cite{dabral18} & 28.0 & 30.7 & 39.1 & 34.4 & 37.1 & 28.9 & 31.2 & 39.3 & 60.6 & 39.3 & 44.8 & 31.1 & 25.3 & 37.8 & 28.4 & 36.3\\
Pavllo \textit{et al.} CVPR’19 \cite{pavllo19} (n=243 CPN) & 34.1 & 36.1 & 34.4 & 37.2 & 36.4 & 42.2 & 34.4 & 33.6 & 45.0 & 52.5 & 37.4 & 33.8 & 37.8 & 25.6 & 27.3 & 36.5\\
Liu \textit{et al.} CVPR’20 \cite{liu20} & 32.3 & 35.2 & 33.3 & 35.8 & 35.9 & 41.5 & 33.2 & 32.7 & 44.6 & 50.9 & 37.0 & 32.4 & 37.0 & 25.2 & 27.2 & 35.6\\
Pavllo \textit{et al.} CVPR’19 \cite{pavllo19} (n=243 GT) & - & - & - & - & - & - & - & - & - & - & - & - & - & - & - & 27.2\\
\hline
Liftformer (n=27 CPN) & 34.1 & 37.6 & 37.1 & 37.5 & 38.6 & 43.5 & 34.1 & 33.7 & 46.7 & 51.4 & 39.5 & 34.0 & 38.8 & 25.3 & 29.8 & 37.6 \\
Liftformer (n=81 CPN) & 32.9 & 36.3. & 35.1 & 35.7 & 37.1 & 41.3 & 32.9 & 32.5 & 44.5 & 50.1 & 36.9 & 33.0 & 36.5 & 25.2 & 26.6 & 35.8\\
Liftformer (n=243 CPN) & 32.2 & 35.2 & 34.3 & 34.8 & 36.1 & 40.5 & 32.7 & 31.4 & 44.9 & 49.7 & 36.2 & 32.8 & 34.8 & 23.9 & 24.9 & 35.0 \\
\hline
Liftformer (n=27 GT)  & 25.4 & 29.2 & 28.6 & 26.2 & 27.6 & 29.0 & 29.1 & 25.1 & 33.4 & 34.6 & 28.1 & 27.4 & 26.3 & 21.3 & 22.6 & 27.6\\
Liftformer (n=81 GT)  & 23.7 & 27.2 & 24.6 & 24.1 & 25.8 & 27.5 & \underline{26.8} & 23.9 & 32.8 & 33.6 & 25.6 & 24.1 & 24.4 & 18.1 & 18.8 & 25.4\\
Liftformer (n=243 GT)  & \textbf{22.3} & \textbf{25.2} & \textbf{22.1} & \textbf{22.0} & \textbf{23.0} & \textbf{24.9} & \textbf{24.5} & \textbf{21.6} & \textbf{29.3} & \textbf{30.3} & \underline{23.7} & \textbf{23.4} & \textbf{22.2} & \textbf{15.6} & \textbf{16.2} & \textbf{23.1} \\
\hline
Liftformer (n=27 GT + WS) & 25.4 & 29.9. & 28.3 & 26.4 & 28.1 & 30.3 & 29.0 & 25.6 & 33.2 & 34.7 & 28.0 & 28.0 & 27.2 & 21.3 & 22.4 & 27.9 \\
Liftformer (n=81 GT + WS) & 24.3 & 28.1 & 26.2 & 24.8 & 25.9 & 28.6 & 27.4 & 24.0 & 33.1 & 34.6 & 26.2 & 25.8 & 24.5 & 19.0 & 19.6 & 26.1 \\
Liftformer (n=243 GT + WS) & \underline{22.7} & \underline{25.3} & \underline{22.7} & \underline{22.4} & \underline{23.2} & \underline{25.6} & \textbf{24.5} & \underline{21.9} & \underline{29.5} & \underline{31.2} & \textbf{23.5} & \underline{23.5} & \underline{22.6} & \underline{16.6} & \underline{16.3} & \underline{23.4} \\
\end{tabular}
\end{adjustbox}
\caption{\textbf{Protocol 2} with P-MPJPE (mm): Reconstruction error on Human3.6M with similarity transformation. CPN - Cascaded Pyramid Network. GT - Ground truth. WS - Weight sharing.}
\label{h36m:p-mpjpe}
\end{table*}

\begin{table*}[tp]
\begin{adjustbox}{width=1\textwidth}
\begin{tabular}{l|ccccccccccccccc|c}
& Dir. & Disc. & Eat & Greet & Phone & Photo & Pose & Purch. & Sit & SitD. & Smoke & Wait & WalkD. & Walk & WalkT. & Avg \\
\hline
Pavllo \textit{et al.} \cite{pavllo19} - Single-frame & 12.8 & 12.6 & 10.3 & 14.2 & 10.2 & 11.3 & 11.8 & 11.3 & 8.2 & 10.2 & 10.3 & 11.3 & 13.1 & 13.4 & 12.9 & 11.6 \\
Pavllo \textit{et al.} \cite{pavllo19} - Temporal & 3.0 & 3.1 & 2.2 & 3.4 & 2.3 & 2.7 & 2.7 & 3.1 & 2.1 & 2.9 & 2.3 & 2.4 & 3.7 & 3.1 & 2.8 & 2.8 \\
\hline
Liftformer (n=27) & 1.6 & 1.7 & 1.3 & 1.9 & 1.3 & 1.5 & 1.6 & 1.8 & 1.0 & 1.4 & 1.2 & 1.5 & 2.2 & 1.9 & 1.7 & 1.6\\
Liftformer (n=81) & 1.4 & 1.5 & 1.1 & 1.7 & 1.1 & 1.4 & 1.4 & 1.7 & 0.9 & 1.3 & 1.1 & 1.2 & 2.0 & 1.7 & 1.4 & 1.4 \\
Liftformer (n=243) & 1.3 & 1.4 & 1.0 & 1.6 & 1.1 & 1.3 & 1.3 & 1.6 & 0.9 & 1.2 & 1.0 & 1.2 & 1.9 & 1.6 & 1.3 & 1.3 \\
\end{tabular}
\end{adjustbox}
\caption{\textbf{Protocol 3} with MPJVE (mm/s): Velocity error over the 3D poses}
\label{h36m:mpjve}
\end{table*}

\begin{figure*}[bp]
    \centering
    \includegraphics[width=\textwidth]{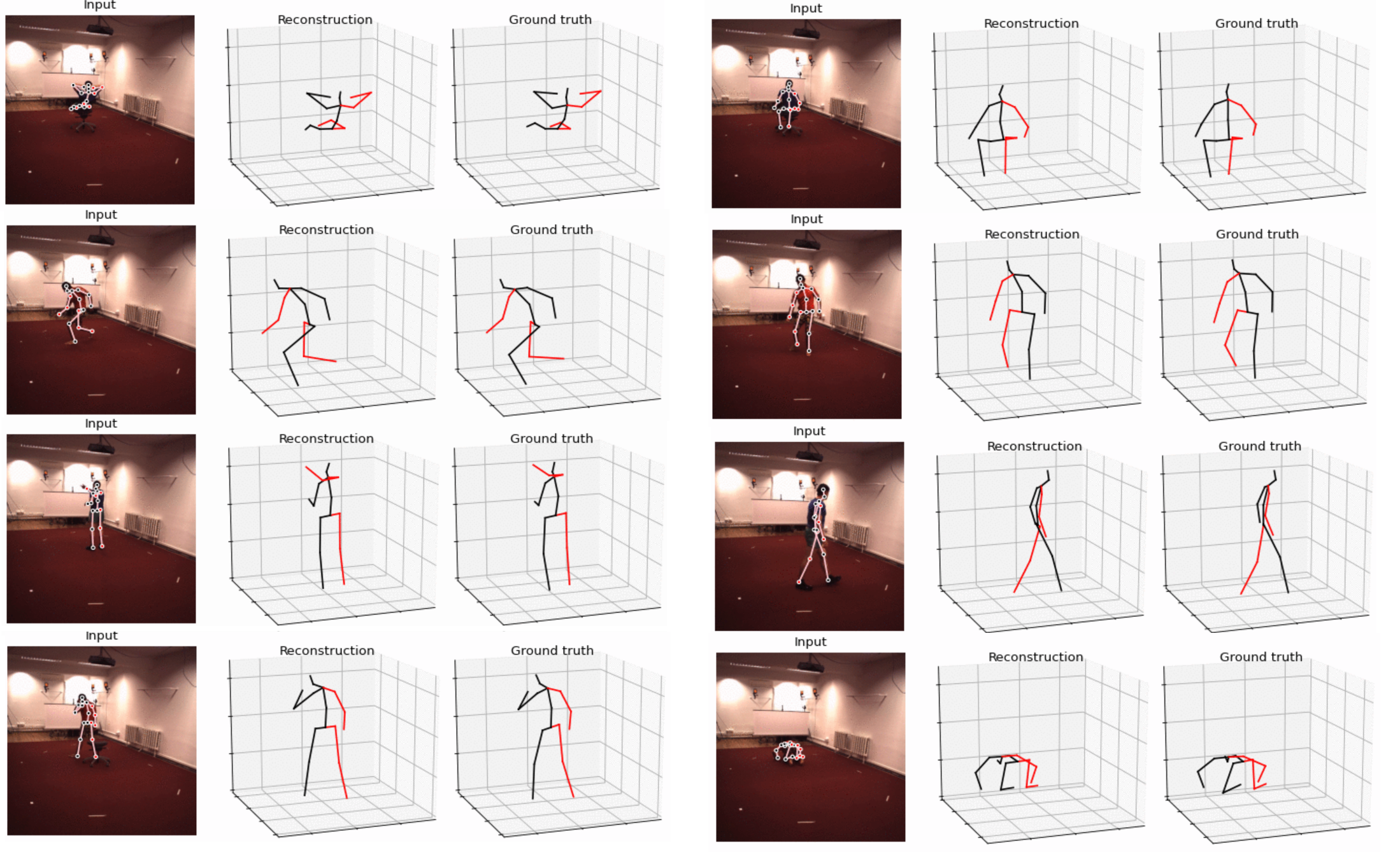}
    \caption{Qualitative results for several actions in the Humans3.6M dataset. From left to right: Original RGB image with 2D keypoint predictions using CPN. 3D reconstruction using LiftFormer (n=243). Ground truth 3D keypoints.}
    \label{fig:qualitative_results}
\end{figure*}

\subsection{Implementation details}
Previous work in \textit{lifting} pose estimation has utilized ground truth human bounding boxes, followed by 2D pose estimators. The most common 2D estimation models are Stacked Hourglass \cite{newell16}, Mask-RCNN \cite{he17} or Cascaded Pyramid Networks\cite{chen18}. Since our approach is detector agnostic, we have investigated several 2D detectors that do not rely on ground truth data. Namely, we have used SH and CPN as detectors for the Human3.6M dataset and Mask-RCNN for HumanEva. However, we have also trained with ground truth 2D poses, to see the reach of our approach.

Following Pavllo \textit{et al.} \cite{pavllo19} methodology, SH  was pretrained on MPII and fine-tunned on Humans3.6M. Both Mask-RCNN and CPN are pretrained on COCO \cite{lin14} and then finetunned on the 2D poses of Human3.6M, since the keypoints in each dataset are defined differently. More specifically, Mask-RCNN uses a ResNet-101 with FPN. Since CPN requires bounding boxes, Mask-RCNN is employed first to detect humans. It then uses a ResNet-50 with \textit{384x384} input resolution. In practice, all 2D predictions have been processed one time first, the results of which are used as input to train our model.

Similarly to the latest work in the field \cite{pavllo19, liu20}, we have divided our tests into 3 sizes of receptive fields (i.e \textit{n=}27, 81, 243).  The only applied data augmentation, for both training and testing, is horizontally flipping the poses. Following the requirements to train the Transformers attention layers, we use the Adam optimizer \cite{reddi18}. All training runs last 80 epochs, for Human3.6M, and 1000 epochs, for HumanEva. We use a \textit{Noam} learning rate schedule which linearly increases for the first training steps \cite{goyal17} (1000 iterations, with a learning rate factor of 12) and then decreases proportionally to the inverse square root of the step number. The batch sizes are proportional to the receptive field value, such that for $b=5120, 3072, 1536$ for $n=27, 81, 243$, respectively. 

In terms of hardware, 8 NVIDIA V100 GPUs have been used to train and evaluate the system, with parallel optimization. Typically, the training time for each receptive field is 8, 14 and 40 hours, respectively, taking the batch sizes into account.

\subsection{Comparison with State-of-the-Art}
\label{ablations}
Tables \ref{h36m:mpjpeCPN} and \ref{h36m:mpjpeGT} show results for \textit{Protocol 1} of our LiftFormer model, trained on both CPN predictions and GT, respectively, for 3 values of receptive fields (27, 81 and 243). Using both CPN and GT predictions as inputs, our model surpasses the current SOTA for all \textit{lifting} monocular methods \cite{liu20} by 0.3\textit{mm} (0.7\%) and 2.8\textit{mm} (8.1\%), respectively. 

Interestingly, weight sharing does indeed produce better results for GT inputs and a receptive field of 243. Also, our model does not rely on additional data, like MPII \cite{pishchulin16} and MPI-INF-3DHP \cite{mehta17}, as some other approaches do. 

\begin{table}[tp]
\begin{adjustbox}{width=1\columnwidth}
\begin{tabular}{l|ccc|ccc|c}
\hline
&  & Walk &  &  & Jog &  & \\
& S1 & S2 & S3 & S1 & S2 & S3 & Avg\\
\hline
Martinez \textit{et al.} ICCV’17 \cite{martinez17} & 19.7 & 17.4 & 46.8 & 26.9 & 18.2 & 18.6 & 24.6\\
Pavlakos \textit{et al.} CVPR’17 \cite{pavlakos18} & 22.3 & 19.5 & 29.7 & 28.9 & 21.9 & 23.8 & 24.4\\
Lee \textit{et al.} ECCV’18 \cite{lee18} & 18.6 & 19.9 & 30.5 & 25.7 & 16.8 & 17.7 & 21.5\\
Liu \textit{et al.} CVPR’20 \cite{liu20} & 13.1 & 9.8 & 26.8 & 16.9 & 12.8 & 13.3 & 15.4\\
Zhou \textit{et al.} ICCV'19 \cite{zhou19} & 13.5 & 9.9 & 17.1 & 24.5 & 14.8 & 14.4 & 15.2 \\
Pavllo \textit{et al.} CVPR’19 \cite{pavllo19} & 13.9 & 10.2 & $46.6^1$ & 20.9 & 13.1 & 13.8 & 14.4 \\
Cheng \textit{et al.} ICCV'19 \cite{cheng19} & 11.7 & 10.1 & 22.8 & 18.7 & 11.4 & 11.0 & 14.3\\
Cheng \textit{et al.} arXiv'20 \cite{cheng20} & 10.6 & 11.8 & 19.3 & 15.8 & 11.5 & 12.2 & 13.5\\
\hline
Liftformer (243 MRCNN) & 17.6 & 11.1 & $48.1^1$ & 30.0 & 14.6 & 15.0 & 25.7 \\
Liftformer (81 MRCNN) & 14.8 & 10.1 & $47.5^1$ & 23.8 & 13.5 & 14.0 & 15.2 \\
Liftformer (27 MRCNN) & 14.0 & 9.3 & $46.6^1$ & 20.1 & 12.6 & 12.8 & 13.8 \\
\hline
Liftformer (243 GT)  & 12.0 & 7.6 & 18.1 & 18.1 & 8.8 & \underline{8.9} & 12.3\\
Liftformer (81 GT)  & \underline{10.6} & \underline{7.3} & \textbf{15.9} & \underline{15.4} & \underline{7.9} & 9.0 & \underline{11.0}\\
Liftformer (27 GT) & \textbf{9.4} & \textbf{6.7} & \underline{16.4} & \textbf{14.0} & \textbf{7.6} & \textbf{8.8} & \textbf{10.5}\\
\end{tabular}
\end{adjustbox}
\caption{\textbf{Protocol 2} with P-MPJPE (mm): Reconstruction error on HumanEva. GT - Ground truth (15 joints). MRCNN - Mask-RCNN (17 joints)  \textbf{Note$^1$ :} The high error on \textit{Walk} of S3 is due to corrupted mocap data, hence the average is not calculated with this value.}
\label{heval:p-mpjpe}
\end{table}

Additionally, when using GTs, LiftFormer is still better than other end-to-end models \cite{cheng19, cheng20, zhou19} with 42.9, 40.1 and 39.9 mm MPJPE, respectively, which not only leverage temporal data, but also features extracted from the original RGB images themselves, optical flow or occlusion enhanced heatmaps. 

Our model also outperforms other SMPL-based approaches like SPIN \cite{kolotouros19} or ENAS \cite{pham20}, with 41.1mm and 42.4mm MPJPE, respectively, and multi-view methods, like DeepFuse \cite{huang20} with 37.5mm MPJPE.

Table \ref{h36m:p-mpjpe} reinforces the results of LiftFormer under \textit{Protocol 2}. When using CPN predictions, our model slightly surpasses all previous \textit{lifting} methods \cite{liu20} (0.6mm, 1.7\%). When using GT, our two best models (with receptive fields of 243 and 81, and no weight sharing) achieve 23.1 and 25.4 mm P-MPJPE, which consistently outperform all previous best models \cite{pavllo19} by 4.1mm (15\%) and 1.8mm (6.6\%), respectively. It is worth noticing how weight sharing decreases the performance slightly in this scenario.

Table \ref{h36m:mpjve} shows that our model reduces the MPJVE of the single-frame and temporal baseline \cite{pavllo19} by 88.8\% and 53.36\%, respectively. This results in more temporally coherent poses with a smoother transition between timesteps.

Table \ref{heval:p-mpjpe} displays results on the HumanEva dataset and shows how our model correctly generalizes to smaller datasets. The performance is comparable to other approaches, even when using MRCNN detections, which have been proven to perform worse than CPN detections. Our model outperforms previous state-of-the-art methods, when using GT 2D keypoints, by 3mm (22.2\%). Interestingly enough, smaller receptive fields (\textit{n}=27) seem to perform better in this scenario, similarly to \cite{liu20}.

Overall, our models outperform the latest SOTA \textit{lifting} approaches in terms of performance, whilst keeping the models size relatively small. For a receptive field of 243 and GT 2D inputs, Pavllo \textit{et al.} \cite{pavllo19} has 16.95M parameters and 37.8 MPJPE:  Liu \textit{et al.} \cite{liu20} has 11.25M and 34.7 MPJPE; and our models have 18.96M and 32.5 MPJPE. The next section will show that even when reducing our models parameters, we still outperform both previous approaches.

\begin{table}[t]
\begin{adjustbox}{width=1\columnwidth}
\begin{tabular}{c|c|c|ccc|c|c}
\begin{tabular}{@{}c@{}}Hidden \\ dimension\end{tabular} & \begin{tabular}{@{}c@{}}Multi \\ head\end{tabular} & Blocks & MPJPE. & P-MPJPE & N-MPJPE & MPJVE & Params.\\
\hline
128 & 8 & 6 & 50.1 & 38.6 & 48.1 & 4.69 & 3.57M\\
256 & 8 & 6 & 49.3. & 37.7 & 47.2 & 3.55 & 7.91M\\
\textbf{512} & 8 & 6 & 48.6 & 37.6 & 46.8 & 2.98 & 18.96M\\
768 & 8 & 6 & 48.6 & 37.4 & 46.5 & 2.92 & 33.15M\\
\hline
512 & 4 & 6 & 48.8 & 37.7 & 47.0 & 3.00 & 18.96M\\
512 & \textbf{8} & 6 & 48.6 & 37.6 & 46.8 & 2.98 & 18.96M\\
512 & 16 & 6 & 48.8 & 37.7 & 46.7 & 3.21 & 18.96M\\
\hline
512 & 8 & 4 & 48.8 & 37.7 & 46.9 & 3.78 & 12.65M\\
512 & 8 & \textbf{6} & 48.6 & 37.6 & 46.8 & 2.98 & 18.96M\\
512 & 8 & 8 & 48.6 & 37.5 & 46.6 & 2.83 & 25.26M\\
\end{tabular}
\end{adjustbox}
\caption{Ablation study on the dimensionality of hidden layers, multi-attention heads and encoder blocks in the Transformer model. Inputs are CPN 2D predictions, with a receptive field of 27 and no weight sharing is used.}
\label{ablat:hid_dim}
\end{table}

\subsection{Ablation studies}
Table \ref{ablat:hid_dim} displays the results using CPN predictions and 27 receptive fields. It shows how increasing the dimensionality of the hidden layers in the Transformers attention leads to an increased accuracy of the \textit{lifted} poses. However, this difference becomes negligible when increasing the dimension value from 512 to 768, at the cost of increasing the models parameters by 75\% (19M to 34M). For this reason, most of the reported results of this work use the value 512.

\begin{table}[t]
\begin{adjustbox}{width=1\columnwidth}
\begin{tabular}{c|c|ccc|c}
\begin{tabular}{@{}c@{}}Receptive \\ field\end{tabular} & \begin{tabular}{@{}c@{}}Weight \\ sharing\end{tabular} & MPJPE & P-MPJPE & N-MPJPE & Params.\\
\hline
27 & \xmark & 37.7 & 27.6 & 34.7 &\\
81 & \xmark & 34.5 & 25.4 & 31.2 & 18.96\\
243 & \xmark & 32.5 & 23.1 & 27.8 &\\
\hline
27 & \cmark & 37.7 & 27.9 & 34.9 &\\
81 & \cmark & 35.3 & 26.1 & 32.3 & 13.7\\
243 & \cmark & 31.9 & 23.4 & 28.5 &\\
\hline
243 (B=4) & \cmark & 33.0 & 24.3 & 29.8 & 9.5\\
\hline
243 (B=2, hd=256) & \cmark & 37.5 & 27.5 & 34.8 & 2.4\\
\end{tabular}
\end{adjustbox}
\caption{Ablation study on the weight sharing of attention layers using GT 2D predictions as inputs. B - number of encoder blocks. hd - hidden dimensionality of Transformer Encoder layers}
\label{ablat:weight_sharing}
\end{table}

One of the multiple benefits of using these self-attention models is that the total number of parameters is independent of the number of multi-attention heads. This is because the linear layers in these heads are applied before the splitting and after the concatenation of the heads, so the number of parameters is not affected by it. Since the number of attention heads does not seem to influence the performance, we stick to a baseline with 8 multi-heads.

Finally, the number of blocks is clearly proportional to the model size. We see how increasing the blocks produces better accuracy, but the improvement is negligible when going from 6 to 8 blocks, which corresponds to a significant increase in the models size.

Table \ref{ablat:weight_sharing} demonstrates how using the weight sharing technique for the multi-attention head drastically reduces the number of parameters. We see how weight shared models, given 2D GT inputs, barely decrease their performance, and in the case of 243 receptive field, it actually improves the MPJPE. 

In particular for \textit{Protocol 1}, Table \ref{h36m:mpjpeGT} shows that for a receptive field of 27 our model obtains the same results with and without weight sharing. For a receptive field of 81 using weight sharing decreases slightly the performance. Finally, for a receptive field of 243, the performance is slightly improved. 

However, for \textit{Protocol 2}, Table \ref{h36m:p-mpjpe} shows that weight sharing decreases slightly the performance for all receptive fields. Thus, weight sharing does not necessarily improve nor deteriorate the performance, but depends on specific cases.

For the sake of reducing parameters, we show how we can make really small models, of 9.5 M and 2.4M parameters which are still competitive, by simply tuning some of the Transformer's hyper-parameters. 

\begin{table}[t]
\begin{adjustbox}{width=1\columnwidth}
\begin{tabular}{l|c|c|c|c}
Method & SH PT & SH FT & CPN FT & GT\\
\hline
Martinez \textit{et al.} \cite{martinez17} & 67.5 & 62.9 & - & 45.5 \\
Hossain \textit{et al.} \cite{hossain18} & - & 58.3 & - & 41.6 \\
Lee \textit{et al.} \cite{lee18} & - & - & - & 38.4 \\
Pavllo \textit{et al.} \cite{pavllo19} & 58.6 & 53.4 & 46.8 & 37.2\\
Liu \textit{et al.} \cite{liu20} & \underline{57.3} & \underline{52.0} & \underline{45.1} & \underline{34.7}\\
Liftformer (n=243) & \textbf{57.0} & \textbf{51.3} & \textbf{44.8} & \textbf{31.9}\\
\end{tabular}
\end{adjustbox}
\caption{Accuracy of 3D predictions in terms of 2D detectors under \textit{Protocol 1} with MPJPE (mm). PT - pre-trained, FT - fine-tuned, GT - Ground truth, SH - stacked hourglass, CPN - cascaded pyramid network.}
\label{ablat:detectors}
\end{table}

For the 9.5M model, we see how we still have a better accuracy and smaller model size than \cite{pavllo19} and \cite{liu20}. They obtain 37.8 and 34.8 MPJPE, respectively, whereas we achieve 33.0 MPJPE. This is with model sizes of 16.95M, 11.25M and 9.5M, respectively. Even for the smallest case (2.4M parameters, 2 encoder blocks, weight sharing and 256 hidden dimensionality), our model is only 5.6mm worse than our best previous result, and we still outperform Pavllo's \textit{et al.} \cite{pavllo19} results in terms of model size (2.4 vs 16.95M) and accuracy (37.5 vs 37.8 MPJPE),

Table \ref{ablat:detectors} displays the results based on the 2D predictors: pre-trained Stacked Hourglass on MPII (PT SH), fine-tuned Stacked Hourglass on Human3.6M (SH FT), fine-tuned Cascaded Pyramid Network (CPN FT) and ground truths (GT). As demonstrated by previous research, the quality of the initial 2D predictions is a clear indication of the final 3D accuracy. More interesting is the fact that our model outperforms all previous \textit{lifting} approaches independently of the 2D detector used.

\section{Conclusions}
We have proposed the usage of a self-attention Transformer Encoder model to estimate the depth of 2D keypoints in monocular videos. The self-attention architecture allows the model to produce temporally coherent poses by exploiting the long-range temporal information across frames/poses.

When using 2D predictions (Mask-RCNN and CPN) as inputs, our model outperforms all previous \textit{lifting} approaches and is comparable to methodologies that use both keypoints and features extracted from the original RGB images. For ground truth inputs, our model outperforms all previous models substantially, achieving results comparable to SMPL or multi-view approaches.

\clearpage
\printbibliography
\end{document}